\documentclass[11pt]{article}

\usepackage[T1]{fontenc}

\usepackage{amssymb}
\usepackage{amsmath}
\usepackage{authblk}
\usepackage{bm}
\usepackage{cite}
\usepackage[dvipsnames]{xcolor}
\usepackage{soul} 
\usepackage{hyperref}

\usepackage[top=2cm, bottom=2cm, left=2cm, right=2cm]{geometry}

\usepackage{graphicx}

\title{Inversion dynamics of class manifolds in deep learning reveals tradeoffs underlying generalisation}

\author[1]{Simone Ciceri}
\author[1]{Lorenzo Cassani}
\author[3]{Matteo Osella}
\author[2]{Pietro Rotondo}
\author[3]{Filippo Valle}
\author[1,2,*]{Marco Gherardi}
\affil[1]{Universit\`a degli Studi di Milano, via Celoria 16, 20133 Milan, Italy}
\affil[2]{Istituto Nazionale di Fisica Nucleare --- Sezione di Milano, via Celoria 16, 20133 Milan, Italy}
\affil[3]{Universit\`a degli Studi di Torino and INFN, Sezione di Torino, via Giuria 1, 10125 Turin, Italy}
\affil[*]{marco.gherardi@unimi.it}

\begin{document}

\maketitle

\abstract{
To achieve near-zero training error in a classification problem,
the layers of a feed-forward network have to disentangle the manifolds of data points with different labels, to facilitate the discrimination.
However, excessive class separation can bring to overfitting since good generalisation requires learning invariant features, 
which involve some level of entanglement.
We report on numerical experiments showing how the optimisation dynamics finds representations that balance these opposing tendencies 
with a non-monotonic trend. After a fast segregation phase, a slower rearrangement (conserved across data sets and architectures) increases the class entanglement.
The training error at the inversion
is stable under subsampling, and across
network initialisations and optimisers, which characterises
it as a property solely of the data structure and (very weakly) of the architecture.
The inversion is the manifestation of tradeoffs elicited by well-defined and maximally stable 
elements of the training set, coined ``stragglers'', particularly influential for generalisation.}

\section*{Introduction}

Supervised deep learning excels in the baffling
task of disentangling the training data, so as to reach
near-zero training error, while still achieving good accuracy
on the classification of unseen data.
How this feat is achieved,
particularly in relation to the geometry and structure of the training data,
is currently a topic of debate and partly still an open question
\cite{
PacelliAriosto:2023,
WakhlooSussman:2023,
PetriniCagnetta:2023,
FengZhang:2023,
BaldassiLauditi:2022,
IngrossoGoldt:2022,
AdvaniSaxeSompolinsky:2020,
GoldtMezard:2020,
Mezard:2017,
Neyshabur2017,
zhang2016understanding,
MartinMahoney2017}.
Activations of hidden layers in response to input examples,
i.e., the internal representations of the data,
evolve during training to facilitate eventual linear separation
in the last layer. This requires a gradual segregation of points
belonging to different classes, in what can be pictured as a
disentangling motion between their class manifolds.

Segregation of class manifolds is a powerful conceptualisation
that informs the design of distance-based losses in metric learning and contrastive learning
\cite{KhoslaTeterwak:2020,
KamnitsasCastro:2018,
HofferAilon:2015,
SalakhutdinovHinton:2007,
ChopraHadsell:2005}
and underlies several approaches aimed at quantifying expressivity and
generalisation, in artificial neural networks as well as in neuroscience
\cite{SchillingMaier:2021,
ChungLeeSompolinsky:2018,
RussoBittner:2018,
KadmonSompolinsky:2016,
PaganUrban:2013,
DiCarloCox:2007}.
Several recent efforts have leveraged this picture
to characterise information processing along the layers of a deep network,
particularly focusing on metrics such as intrinsic dimension and curvature
\cite{FarrellRecanatesi:2022,
CohenChung:2020,
AnsuiniLaio:2019,
FarrellRecanatesi:2019,
RecanatesiFarrell:2019,
PooleLahiri:2016}.
In Ref.~\cite{CohenChung:2020}, for instance,
two descriptors of manifold geometry, related to the intrinsic dimension
and to the extension of the manifolds, are shown to undergo dramatic reduction
as a result of training in deep convolutional neural networks.
Such shrinking decisively supports the model's capacity in a memorisation task.

Yet, this appears to be just one side of the coin.
There are indications that entanglement
of class manifolds in the internal representations of deep neural networks promotes the correct discrimination of test data
\cite{FrosstPapernot:2019}.
This fact appears counterintuitive,
as more entangled representations should correspond
to smaller margins. 
Still, manifold entanglement may encourage compression
(in information-theoretic, rather than geometric, meaning)
by reducing the number of discriminative features
and by minimising the information about the input data that gets
propagated through the network,
effectively acting as a regularisation
\cite{AchilleSoatto:2019,
AchilleSoatto:2018,
ShwartzzivTishby:2017
}.

What emerges is a competition between
learning invariant features and disentangling explanatory factors
\cite{Bengio:2013}.
In this perspective, the classic bias-variance tradeoff, and the tension between train and test
accuracy, translate to opposing tendencies for the optimisation dynamics:
segregation of class manifolds on the one hand, and their entanglement on the other.
How this tradeoff is realised dynamically through training
is the focus of this manuscript.

In the spirit of statistical physics \cite{Zdeborova:understanding},
we explore these questions in simple models,
where patterns are more likely to emerge clearly, and
exploration of their causes and consequences is less hampered
by confounding factors.
As an illustrative example, we consider
 a two-layer fully connected network, using
$P=8192$ points, $\left\{x^\mu\right\}$, from MNIST,
a dataset widely employed in computer vision,
containing $28\times 28$ greyscale images of handwritten digits.
We train the network to solve the parity classification
task, where the label is $+1$ for even digits
and $-1$ for odd ones.
However, we anticipate that the phenomenology we will describe using this simple setting is more general:
it is present also when training wider and deeper networks, as well as in more challenging data sets,
such as KMNIST and CIFAR-10, and for different classification tasks (see Results and Supplementary Information).

At each epoch $t$ during training, the activation of the hidden layer
is a function $h_t$ mapping elements of $\mathbb R^N$
to elements of $\mathbb R^H$, where $N$ is the dimension of the input space
and $H$ is the width of the hidden layer.
Our goal is to observe the evolution, throughout training,
of the internal representations $h_t(x^\mu)$ of the training data
$x^\mu\in\mathcal T$.
In particular, we focus on the overall dispersion
of points belonging to the same class $y$, i.e., of 
the images under $h_t$ of equally-labelled elements of the training set.
The projective nature of linear separability
suggests to consider projections onto the unit sphere $\mathcal S^{n-1}$:
$\hat h_t(x^\mu)= h_t(x^\mu)/\lVert h_t(x^\mu) \rVert$
(see the Methods for further explanation).
Such normalisation is natural when $h_t$ is the representation at the last layer,
but we employ the same definition even when considering the first layer
in a deep network.
Thus, the internal representations of the two classes,
or ``class manifolds'', at each epoch $t$, are the two sets
\begin{equation}\label{eq:manifolds}
\mathcal M_{\pm}(t) = \{\hat h_t(x^\mu) \;|\; y(x^\mu)=\pm 1\}
\subset \mathbb R^H,
\end{equation}
where $y(x^\mu)$ is the label of $x^\mu$.
Intuitively, separation of the two classes by the last layer is facilitated
whenever $\mathcal M_+(T)$ and $\mathcal M_-(T)$, at the final epoch $T$,
are small or far apart.
This intuition is confirmed by analytical computations
\cite{Gherardi:2021,ChungLeeSompolinsky:2018}.

Our analysis is based on a simple descriptor of manifold extension, the gyration radius,
a metric proxy of the set's extension in Euclidean space.
The two radii $R_\pm(t)$,
together with the distance $D(t)$ between the two centres of mass
of the two sets $\mathcal M_\pm(t)$,
are three metric quantities recapitulating
the geometry of the internal representations
$\left\{h_t(x^\mu)\right\}$
(see Methods).
%
%

\section*{Results}

\subsection*{Class manifold segregation in shallow networks}

The internal representations
of data points belonging to the same class
are expected to move closer to one another after training.
This is persuasively shown in \cite{CohenChung:2020}, where the authors focus on state-of-the-art models
(AlexNet and VGG-16) and on a sophisticated data set such as ImageNet.
It is not obvious whether the systematic compaction
that they observe is due specifically to
special properties of the complex ImageNet data set,
or to the heavily convolutional architectures probed.
To address these questions, we trained
a shallow network on the simple task described above,
and compared $R_\pm$ and $D$ before and after training.
Figure 1(a) shows that the internal representations $\mathcal M_\pm$
in the trained model are always less entangled than at initialisation, i.e, 
they are more compact ($R_\pm$ is smaller) and further apart ($D$ is larger).

\begin{figure}[t]
\centering
\includegraphics[width=\textwidth]{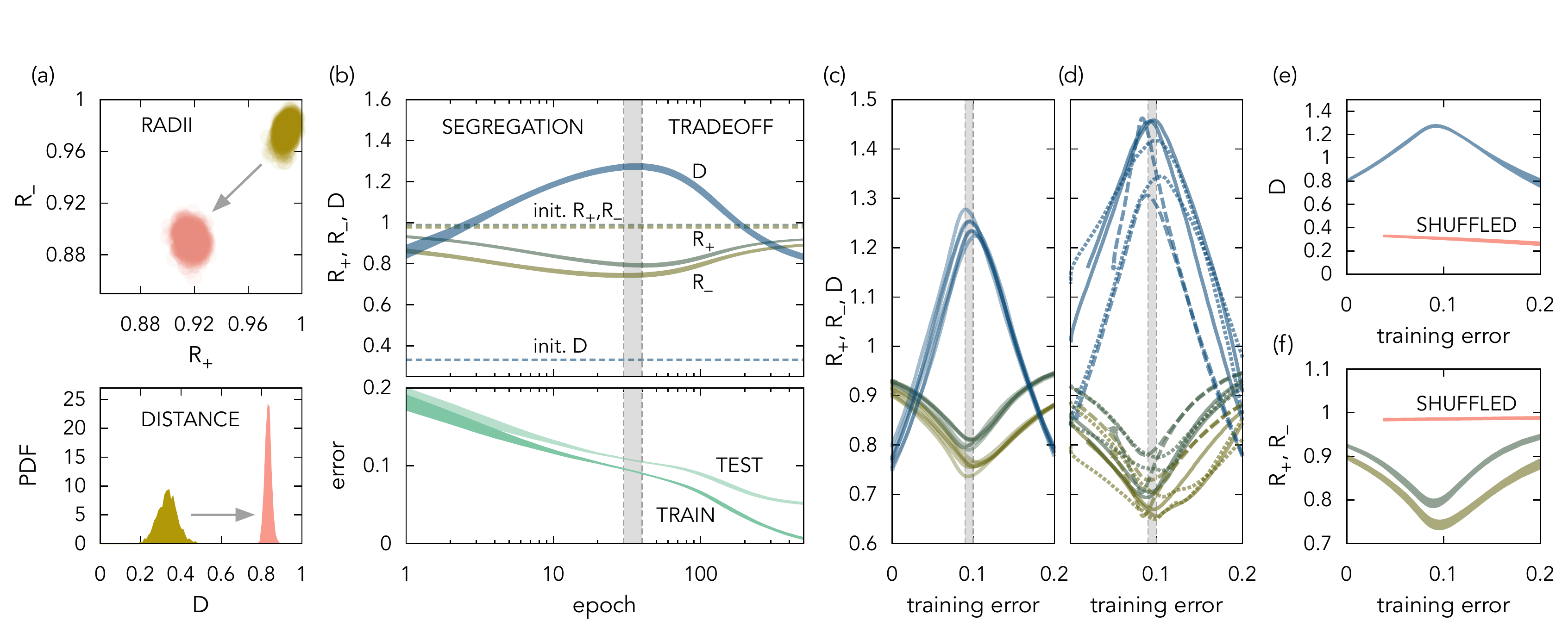}
\caption{
\footnotesize{
\textbf{Non-monotonic learning dynamics.}
(a) Training disentangles the class manifolds.
Scatter plot of the two radii $R_\pm$ (top) and histograms of the distance $D$ (bottom)
from 1000 independent runs, at initialization (yellow) and after training 
(pink).
(b) Class manifold dynamics is non-monotonic.
Radii and distance (top) and train and test errors (bottom)
as functions of training epoch (on the x axis, in log scale);
the dashed horizontal lines are the mean values at initialisation;
inversion happens in the grey shaded regions;
curve widths are 2 standard deviations.
(c)
Dynamics is robust to sub-sampling.
The three metric quantities as functions of training error
(only means shown, computed over 20 runs);
different curves are obtained by training on non-overlapping
subsets of MNIST.
(d)
Dynamics is similar across optimisers and hyperparameters.
Solid lines: Adam (learning rates $0.001$ and $0.005$);
dashed lines: GD with weight decay ($\lambda=0.01$ and $0.05$);
dotted lines: GD with momentum ($\mu,\eta=0.5,0.5$ and $0.9,0.2$).
Curves are averages over 20 runs.
(e,f)
Randomised labels (pink curves) remove the non-monotonicity.
}}
\end{figure}

\subsection*{Dynamics of class manifolds is non-monotonic}

By contemplating the temporal dimension as well,
one can address questions regarding the dynamics of manifold segregation.
In particular, do the metric quantities evolve monotonically?
Figure 1(b) shows that the answer is negative:
$R_\pm$ and $D$ significantly overshoot before converging to their asymptotic
values.
An ``inversion epoch'' $t_*$ marks the separation between two qualitatively
different training periods.
During the first, which happens fairly quickly,
the internal representations of points belonging
to the same class are brought closer to one another,
while the representations of points belonging to different classes
move further away from each other.
After $t_*$, when the radii $R_\pm$ stop decreasing
and the distance $D$ stops increasing, training proceeds by
a slow expansion of the manifolds and a gradual drift of their centres of mass,
bringing them closer together.
During the latter ``expansion'' phase, neither the radii nor the distance get back
to their pre-training values.

\subsection*{Invariance of the training error at the inversion point}

The location of the inversion epoch $t_*$ where the 
time derivatives of the metric quantities change sign
(i) is not appreciably different between $R_+$, $R_-$, and $D$
(we obtain $t_*\approx$ 30--40, corresponding to the grey area in Fig.~1);
(ii) it barely fluctuates between runs started from different initialisations;
(iii) it is not a special point for either the test or the training errors
(Fig.~1(b));
see the Methods for definitions.

Since the inversion is an intrinsically dynamical phenomenon,
an important question is how it depends on the optimisation dynamics.
To address this question, we considered the metric quantities
as functions of the training error $\epsilon_\mathrm{tr}$, by computing them from the sets
$\mathcal M_\pm(t(\epsilon_\mathrm{tr}))$.
The epoch $t(\epsilon_\mathrm{tr})$ here is defined as the first epoch
when the training error crosses the value $\epsilon_\mathrm{tr}$.
Figure 1(d) shows that, although $t_*$ itself can be
very different for different optimisers, the training error at the inversion epoch
$\epsilon_\mathrm{tr}(t_*)$ is approximately invariant;
the figure collects trajectories obtained by running Adam and
gradient descent (GD) with different learning rates, with and without momentum and weight decay.
Using stochastic GD with a batch size much smaller than the training set
gives similar results, but shifts the inversion slightly towards smaller
training errors
(Supplementary Figure 1).

Similarly, we can ask whether the training error at the inversion is sensitive to 
sampling noise in the training data.
Figure 1(c) shows that the dynamics, and $\epsilon_\mathrm{tr}(t_*)$ in particular,
is quite independent of the specific subset of the training set employed for training.

\subsection*{Manifold expansion is elicited by structure in the data}

What is causing the expansion phase?
We will give an answer in the upcoming sections.
A preparatory question is the following:
does the inversion dynamics persist if one destroys
the dependences between the data points and their labels?
We repeated the same experiments
as above, this time with randomly chosen labels for each input.
The expansion phase disappears (Fig.~1(e,f)),
giving way to a single slow segregation mode:
the distance between the latent manifolds increases monotonically,
while the two radii remain roughly constant throughout training.
This result suggests that the non-monotonic dynamics
is elicited by data structure,
i.e., by the relation between the geometry of the data manifolds
and the labels
\cite{Mezard:2023, GoldtMezard:2020, CountingPRR:2020, ChungLeeSompolinsky:2018}.

\begin{figure}[t]
\centering
\includegraphics[width=\textwidth]{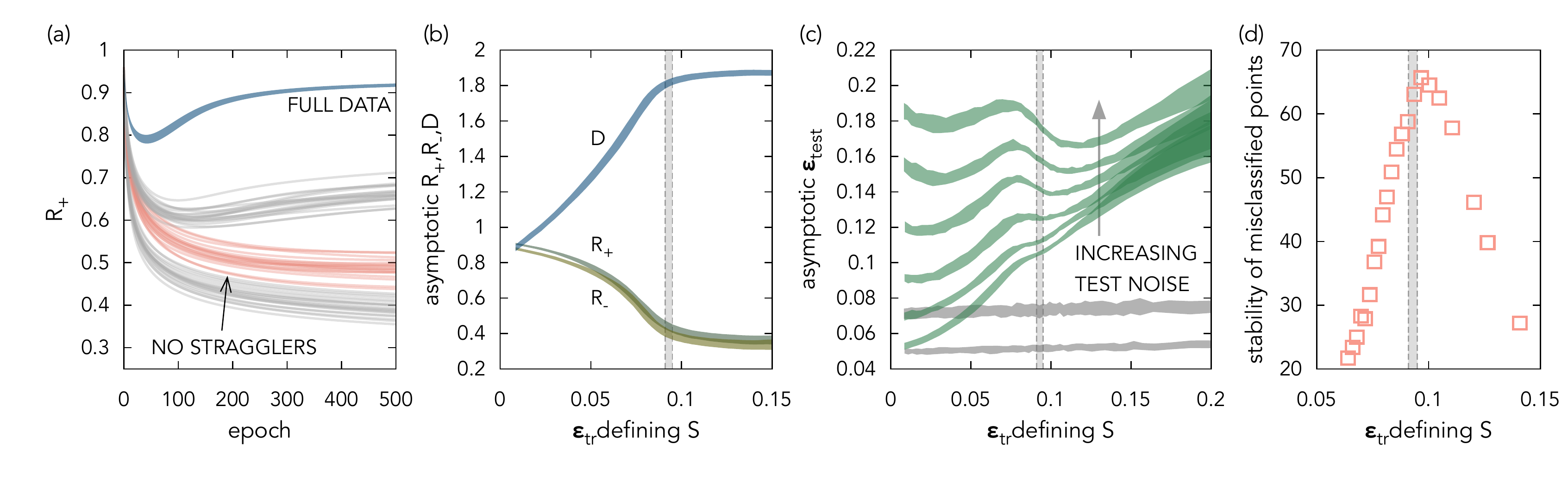}
\caption{
\footnotesize{
\textbf{Stragglers shape the dynamics and influence generalisation.}
(a) Training without stragglers removes the inversion.
The blue curve is obtained by training with the full dataset
(shaded region corresponds to 2 sigmas);
pink curves (indicated by the arrow) are 20 runs
with the pruned training set $\mathcal T\setminus \mathcal S(t_*)$;
the variability is due to the different initialisations, which affect
both the dynamics and the elements of $\mathcal S(t)$;
grey curves above and below the pink ones are obtained with pruned training sets
$\mathcal T\setminus \mathcal S(t)$, with 
$t=100>t_*$ and $t=10<t_*$ respectively.
(b) Metric quantities at convergence (y axis)
using training sets $\mathcal T\setminus\mathcal S\left(t\left(\epsilon_\mathrm{tr}\right)\right)$,
as functions of $\epsilon_\mathrm{tr}$ (x axis).
(c) Removal of stragglers affects the test error at convergence (y axis).
The green curves, from bottom to top, are obtained from noisy test sets,
obtained by adding white noise, independently to each pixel, with standard deviation
$\sigma = 0, 0.5, 0.75, 1., 1.2, 1.5$ respectively
(inputs are standardised, see Methods);
shaded regions correspond to 2 sigmas.
Grey curves are obtained by removing, for each $\epsilon_\mathrm{tr}$,
a random set of points, of the same cardinality as $\mathcal S\left(t\left(\epsilon_\mathrm{tr}\right)\right)$
(only the two smallest values of $\sigma$ are shown).
(d) The inversion point marks a maximally stable set of misclassified points.
Pink crosses are z-scores of the
stability of the set $\mathcal S\left(t\left(\epsilon_\mathrm{tr}\right)\right)$ (y axis; see Methods)
under fluctuations in the initialisations, as a function of $\epsilon_\mathrm{tr}$.
In all plots, $\mathcal T$ contains $P=8192$ elements from MNIST,
the architecture is a two-layer network with 20 hidden units.
}}
\end{figure}

\subsection*{Non-monotonic dynamics reveals trade-offs due to stragglers}

What happens at the inversion?
Some insight can be gained by watching
which subset of the training set is still classified incorrectly at $t_*$.
At $t_*$, the model classifies correctly
most of the training set.
Further optimisation of the loss function
requires to trade off the overall segregation of this bulk
for the separability of the few data points that are still misclassified.

Consider the set of misclassified points at epoch $t$:
\begin{equation}
\mathcal S(t) = \{x^\mu\in\mathcal T \;|\; \hat y_t(x^\mu)\neq y^\mu \},
\end{equation}
where $\hat y_t(x^\mu)$ is the label predicted by the network trained up to
epoch $t$, Eq.~(\ref{eq:predictor}).
We name ``stragglers'' the elements of $\mathcal S(t_*)$,
owing to their being late to catch up with the rest of the training set.

Does the expansion period persist if we remove the stragglers from the training set?
Figure 2(a) shows how $R_+$ behaves when retraining the network on the reduced training set $\mathcal T \setminus \mathcal S(t_*)$.
Removal of $\mathcal S(t_*)$
completely deleted the expansion period,
in favour of a longer, and more seamless, segregation phase.
Not only did the radius decrease monotonically on average:
no inversion point could be identified in any single training run.
Instead, removal of a random subset of the same cardinality
did not affect the radii appreciably.
Pruning the dataset by removing sets $\mathcal S(t<t_*)$,
which are generally larger than $\mathcal S(t_*)$,
had the same effect, but $t_*$ was the largest epoch at which this happened:
removing sets $\mathcal S(t>t_*)$ did not destroy the non-monotonicity
(bottom grey lines in Fig.~2(a)).
Note that the sets $\mathcal S(t)$ depend on the initialisation;
all statements made here were checked for 20 different initialisations.

The individuality of the inversion epoch $t_*$ is emphasised by yet another experiment.
We used the pruned dataset $\mathcal T\setminus\mathcal S\left(t\left(\epsilon_\mathrm{tr}\right)\right)$,
and measured the metric quantities at convergence, as functions of $\epsilon_\mathrm{tr}$
(Fig.~2(b)).
The training error at the inversion, $\epsilon_\mathrm{tr}(t_*)$, marks the
boundary between two qualitatively different phases:
when $\epsilon_\mathrm{tr}$ 
is larger than $\epsilon_\mathrm{tr}(t_*)$
the asymptotic geometry of class manifolds 
is approximately independent of $\epsilon_\mathrm{tr}$.

In MNIST, with $P=8192$ and a two-layer network with 20 hidden units, 
the number of stragglers is
$\left|\mathcal S(t_*)\right| \approx 800$
(how this number changes for different tasks and architectures
is reported below).
Similarly to the inversion epoch $t_*$, the identity of the stragglers 
is conserved across network initialisations
and, when training with stochastic GD, 
for different shuffles of the training set;
we checked this by comparison with a null hypergeometric model
(see Methods).
Remarkably, among all the sets $\mathcal S\left(t\left(\epsilon_\mathrm{tr}\right)\right)$,
stragglers are maximally conserved (Fig.~2(d)).

\subsection*{Stragglers influence generalisation and noise robustness}

The experiments above elucidated how stragglers
shape the dynamics of the class manifolds, by 
triggering a tradeoff phase where entanglement between different classes,
as measured via their metric properties, increases.
As mentioned in the Introduction,
entanglement between class manifolds is expected, in turn, to facilitate good generalisation,
because it promotes learning of common invariant features.
Persuasive clues that this is in fact the case
were identified in Ref.~\cite{FrosstPapernot:2019}.
Hence, is it possible to quantify the influence of stragglers on generalisation?

We trained a two-layer network
on the pruned training sets
$\mathcal T\setminus\mathcal S\left(t\left(\epsilon_\mathrm{tr}\right)\right)$,
and measured the test error at convergence, $\epsilon_\mathrm{test}$.
The resulting $\epsilon_\mathrm{test}$ as a function of $\epsilon_\mathrm{tr}$,
from which the training set depends, is shown in Fig.~2(c).
Testing accuracy deteriorates ($\epsilon_\mathrm{test}$ increases)
when removing any subset 
$\mathcal S\left(t\left(\epsilon_\mathrm{tr}\right)\right)$ from $\mathcal T$.
The magnitude of the deterioration is much larger than that obtained
when removing a random subset of $\mathcal T$ of the same size as $\mathcal S\left(t\left(\epsilon_\mathrm{tr}\right)\right)$ (grey regions in Fig.~2(c)).

The magnitude of the increase in $\epsilon_\mathrm{test}$
is not a featureless function of $\epsilon_\mathrm{tr}$:
the training error at the inversion epoch
(grey vertical band in the figure) 
appears to separate two different branches of the curve.
This contrast was accentuated when we repeated the same experiment
with noisy versions of the test set, obtained by adding
white noise to the input images with increasingly large variances.
At low signal-to-noise ratios (large variances), the curves become non-monotonic,
signalling a complex relation between the pruned subsets
and the testing accuracy.
Interestingly,
at sufficiently low signal-to-noise ratios, removing stragglers
reduces the test error to values below the original ones.

Given their impact on generalisation, it is natural to ask whether
the role of stragglers is different on the two sides of the double-descent curve
\cite{Belkin:2019,Nakkiran:2021}.
We repeated the preceding experiments with training set sizes and number of learnable
parameters well above and well below the interpolation threshold,
finding no appreciable differences
(Supplementary Figure 2).

\begin{figure}[t]
\centering
\includegraphics[width=\textwidth]{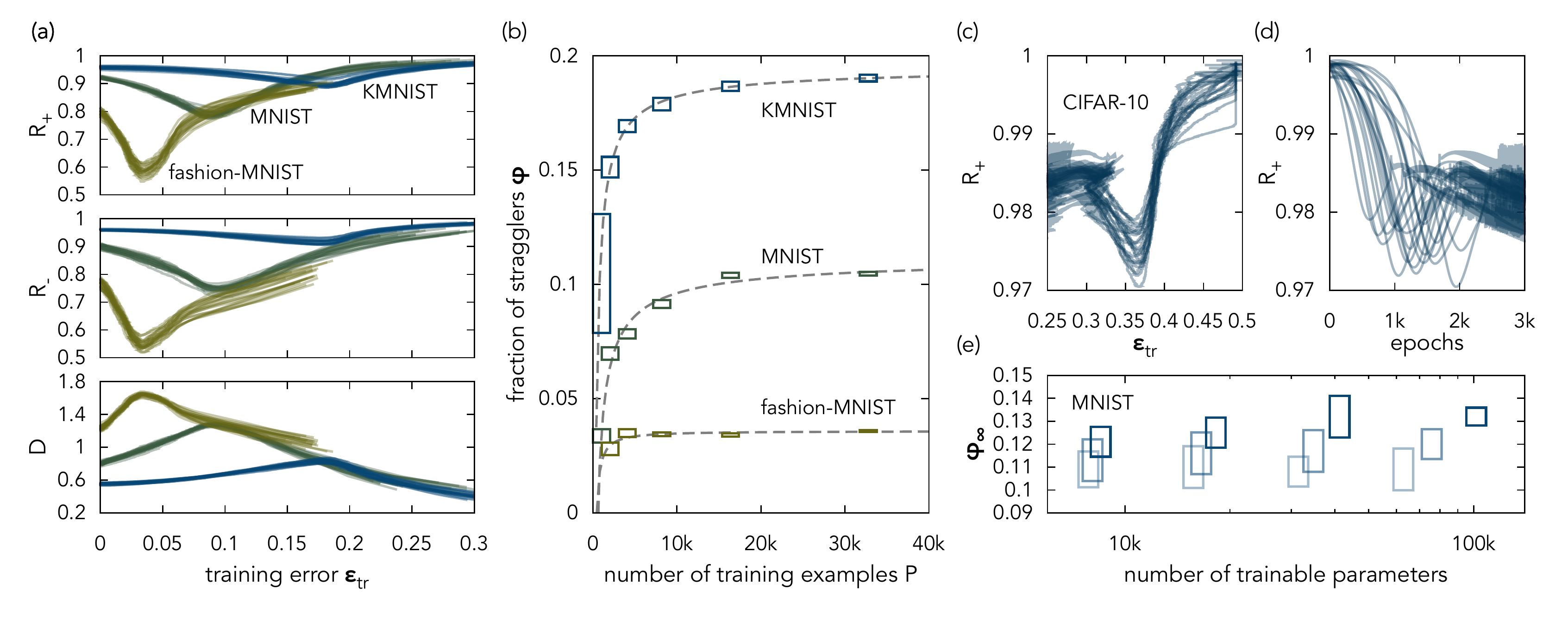}
\caption{
\footnotesize{
\textbf{Stragglers across data sets and architectures.}
(a)
The three metric quantities (y axes) as functions of the
training error (x axis) for MNIST, KMNIST, and fashion MNIST.
(b) Fraction of stragglers has a well-defined large-dataset limit.
Dashed lines are fits of Eq.~(\ref{eq:fss}) to these data.
(c), (d) Non-monotonic dynamics of $R_+$ (y axes) in CIFAR-10,
as a function of training error in (c) and epochs in (d).
(e) The asymptotic (large-dataset) fraction of stragglers (y axis)
depends only weakly on the depth, and negligibly on the width,
of the architecture.
The four groups of boxes correspond to increasing widths
from left to right; darker shades of blue correspond to deeper
architectures.
The curves in (a),(c), and (d) are 20 runs for each data set.
Box heights in (b) and (e) correspond to 2 standard deviations.
Architectures and parameters:
2 layers with 20 hidden units each in (a) and (b);
8 layers (fully connected) with 20 hidden units each, 
learning rate $\eta=0.02$, in (c) and (d);
2,4, and 8 layers, each with 10,20,40, and 80 hidden units,
$\eta=0.1$, in (e).
}}
\end{figure}

\subsection*{Non-monotonic dynamics and stragglers in other data sets}

The non-monotonic segregation dynamics, as discussed above, is due to
data structure.
Is this a peculiarity of MNIST, or is the phenomenology more general?
Figure 3(a) shows that KMNIST and fashion-MNIST, other commonly-employed
data sets (see Methods), engender a similar non-monotonic dynamics.
In addition, pruning the training set has similar consequences to those observed in MNIST.
Stragglers, the misclassified examples at the inversion point,
are again the most conserved among all subsets $\mathcal S(t)$
(Supplementary Figure 3).

The more complex data set CIFAR-10,
which required a more expressive network (see the caption to Fig.~3),
allows us to make an interesting observation.
The goal of defining the inversion point as a function of training error
(as opposed to the epoch) was to enable a fair comparison
between optimisers. For simple data sets such as MNIST,
plotting $R_\pm$ and $D$ as functions of epoch or
training error has no qualitative impact on the observed behavior.
On the contrary, when an 8-layer architecture is trained on CIFAR-10,
the dependence on
the training error is much sharper (less fluctuating).
Figures 3(c) and 3(d) show a comparison between the use of
epochs and training error as independent variables.

\subsection*{Weak dependence on data size,
depth, width, activation function}

The training error at the inversion epoch
$\phi = \epsilon_\mathrm{tr}(t_*)$
is the fraction of stragglers in the data set.
Above, we have used a fixed number of training examples $P$.
How does $\phi$ depend on this choice?
Increasing $P$ makes the training more and more
difficult by adding new constraints in the optimisation problem, thus potentially also influencing the inversion point.
However, this is not the case.
Figure 3(b) shows that $\phi$, as a function of $P$, saturates to a relatively small fraction
for MNIST, KMNIST, and fashion MNIST.
By fitting a tentative scaling form for $\phi$ as a function of $P$,
one can attempt an extrapolation to infinite data set size,
thus obtaining an estimate of the asymptotic fraction $\phi_\infty$
(see Methods, Eq.~(\ref{eq:fss})).
The fitted curves are in Fig.~3(b); we obtained
$\phi_\infty \approx$ $4\%$ (fashion MNIST), $11\%$ (MNIST),
$20\%$ (KMNIST).

The arrangement of these values and the higher inversion point for CIFAR-10 (Fig.~3(c)) indicate a relation between 
the complexity of the data set and the proportion
of stragglers.
A similar relation was reported between the fraction
of critical samples, defined via the concept of adversarial examples, and data-set complexity~\cite{Arpit:2017}.

We explored the dependence of the fraction of stragglers
on the architecture by computing $\phi_\infty$ for
fully connected networks with 2,4, and 8 layers, and 10,20,40, and 80
units per hidden layer.
Figure 3(e) shows $\phi_\infty$ as a function
of the total number of trainable parameters.
In MNIST, about 11--13\% of the training set is composed of stragglers,
this figure being approximately constant over the range of depths and widths considered,
encompassing more than an order of magnitude in total number of parameters.
A weak systematic dependence emerges, mainly as a function of depth. 

We checked the stability of the stragglers' identity across architectures,
by comparing the sets $\mathcal S(t_*)$ obtained in models with different widths and depths.
Stragglers are strongly conserved, with $z$-scores lying close to those obtained
by comparing different training runs of a single shallow network
(see Methods).

Finally, we observed that the inversion point, 
when using nonlinearities other than tanh,
is slightly more fluctuating, but it still occurs around the same
value of the training error.
For a 4-layer network with 20 hidden units per layer, trained on
8192 examples from MNIST, we found
$\phi=0.089\pm 0.009$ (reLU),
$\phi=0.088\pm 0.007$ (leaky reLU with negative slope 0.1),
and $\phi=0.097\pm 0.014$ (siLU),
to be compared with $\phi=0.098\pm 0.002$ (tanh).
The phenomenology persists in
the fully linear case where the activation function is the identity,
for which we found $\phi=0.100\pm0.002$.
This suggests that theoretical insight into the segregation dynamics
of class manifolds may be gained by employing the theory of
deep linear networks, which allows for analytical computations
\cite{SaxeMcclelland:2014}.

\subsection*{Specificity of stragglers within the data set}

Do stragglers occupy special places with respect to the data manifold?
While visual inspection does not reveal striking peculiarities,
we found that stragglers, compared to other training examples,
are significantly further away from the respective class centres.
This analysis was done using
all 10 MNIST classes, even though the classification problem is binary.
By embedding the data in 2-dimensional space by t-SNE,
it becomes visually clear that stragglers lie preferentially close to the class boundaries
(see Supplementary Figure 4).
This result suggests that it may be useful to think of stragglers as
the ``support vectors’’ for the non-linear classification problem.

\section*{Discussion}

The nonmonotonic dynamics, and its inversion point in terms of training error,
proved to be remarkably robust to changes in the hyperparameters and to perturbations.
The fraction of stragglers 
appears to be an invariant property of the data set,
characterising its complexity in terms of the tradeoffs discussed above.
How this measure relates to other metrics of task difficulty,
such as the intrinsic dimensions of the data set
\cite{ErbaGherardi:2019,FaccoDErrico:2017}
or of the objective landscape \cite{LiFarkhoor:2018}, 
and to other specifics of data structure
\cite{RotondoPastore:2020:PRL,
PastoreRotondo:2020:PRE,Arpit:2017,
GherardiRotondo:2016},
is an open question.
In spite of the robustness presented above,
we were able to find one way to disrupt the behaviour.
Increasing the variance of the weight initialisation kept $\phi$
unchanged but pushed the radii towards 1 and made the minimum shallower.
When the variance far exceeded the inverse
of the number of units in hidden layers, the minimum disappeared abruptly
and the radii became monotonic.
This may be the manifestation of a transition between
the feature learning and the lazy training regimes
\cite{GeigerSpigler:2020}.

We list here some limitations of our work.
(i) However robust, the phenomenology found in small fully-connected architectures
should not be expected to arise immediately, or to be as clearcut, in state-of-the-art
deep convolutional neural networks or transformers.
(ii)
We focussed solely on the internal representations at the first layer,
even in deeper architectures.
The dynamics in immediately downstream layers is not dissimilar,
but representations closer to the output display different patterns.
(iii) It is not evident how much the behaviour of the test error under pruning,
Fig.~2(c), is sensitive to the choice of architecture.
A more systematic exploration of these matters is left for future work.
To address, at least partially, the limitations (i) and (iii),
we have explored a set of 2-layers convolutional neural networks (CNN).
Their behaviour varies with the choice of hyperparameters.
Some architectures (e.g., with large kernel sizes) behave in the same
way as fully connected ones, while in others
the dynamics is qualitatively the same only for $D$
(see the Supplementary Figure 5
for an example with $4\times 4$ kernels, on MNIST).
Even in the latter architecture, removal of stragglers 
(as identified with either a CNN or a FC network)
has a profound impact on the training dynamics, 
making the inversion less marked.
The results of
Fig.~2(c), and 2(d) are also still valid
(see Supplementary Figure 6).

Our empirical results shed light also on separate questions
regarding the role of different examples during the training.
Once architecture, optimiser, and training objective are fixed,
thus establishing implicit inductive biases,
the ability to generalise to unseen, possibly out-of-distribution,
data is acquired by relying solely on the training set.
Do all training data coherently cooperate in maximising train and test accuracy? 
Or does heterogeneity,
a well documented feature of empirical data sets
\cite{Mazzolini:2018:PRX,MazzoliniGrilli:2018}, play a role?
The inversion dynamics presented here indicates that two 
different compartments of $\mathcal T$
are involved in shaping distinct periods of training, and appear to
have distinct contributions to generalisation.
Stragglers emerge as a set of challenging instances located at the outskirts of class manifolds, which are memorised during later stages of training.  We hypothesize that they carry information about the geometry of the data distribution, thereby contributing to the fine-grained properties of the learned discrimination boundaries. Omitting these examples from the training set results in more compact class representations but at the cost of decreased generalization.  Conversely, when  the geometric details of the data distribution are blurred by the presence of noise, as in our experiments with noisy test sets (Fig.~2c and Supplementary Fig.~2), removing the stragglers can instead enhance the out-of-distribution generalization.

Previous  literature supports the observation that  training examples are consistently classified at different learning stages, implying the existence of easy and hard examples ~\cite{Arpit:2017}.
Accurate metrics for assessing example difficulty are essential for designing data set pruning strategies~\cite{Sorscher:2022}  and curriculum learning protocols~\cite{bengio2009curriculum}.
Exploring the potential role of stragglers as challenging examples in these contexts is an avenue for future research.


\section*{Methods}

\paragraph{Models and training}

Most of our analysis was carried out on a shallow network with weights
$w \in \mathbb{R}^{H\times N}$ and $v \in \mathbb{R}^{2\times H}$, 
and biases $b\in\mathbb R^2$ and $c\in\mathbb R^H$.
We denote $w_{ij}$, $v_i^a$, $b^a$, and $c_i$
the elements of these vectors, where
$i=1,\ldots,H$; $j=1,\ldots,N$; $a=\pm 1$.
The forward function is
\begin{equation}
f^a(x) = 
\sum_{i=1}^H v^a_i \left[h\left(x^\mu\right)\right]_i + b^a,
\quad a=\pm 1,
\end{equation}
where the vector $h(x^\mu)$
is the internal representation of $x^\mu$; 
its components are
\begin{equation}
\left[h(x^\mu)\right]_i=\sigma\left(
\sum_{j=1}^N w_{ij} x^\mu_j + c_i
\right).
\end{equation}
The transfer function $\sigma$ was tanh for most of our analyisis.
With these definitions, the predictor is
\begin{equation}\label{eq:predictor}
\hat y(x) = \underset{a}{\operatorname{argmax}}
\left(\left\{f^a(x)\right\}_a\right).
\end{equation}
We use the subscript $t$, as in $h_t$ or $\hat y_t$, to specify that
weights and biases are those evaluated at epoch $t$ during training.

All models were trained using full-batch gradient descent
with learning rate $\eta=0.2$ (except where stated otherwise),
with loss function
\begin{equation}
L = -\sum_{\mu=1}^P \log \left[ \mathrm{softmax} \left\{f^a(x)\right\}_a\right]_{y(x^\mu)},
\end{equation}
where $y(x^\mu)$ is the label of $x^\mu$ in the training set.
Weights and biases were initialised as independent random variables
with the uniform distribution $\mathcal U\left(-1/\sqrt{n}, 1/\sqrt{n}\right)$,
where $n$ is the number of weights in the layer.
Using other initialisation schemes, such as He
or Xavier initialisation, does not change the results presented above,
but see the comment regarding initialisation in the Discussion.

\paragraph{Datasets and standardisation}

We used the following data sets: 
\begin{itemize}
\item MNIST, handwritten digits, 28x28 greyscale images \cite{MNIST},
\item Kuzushiji-MNIST, or KMNIST;
cursive Japanese characters, 28x28 greyscale images \cite{KMNIST},
\item 
Fashion MNIST;
Zalando's article images,
28x28 greyscale images \cite{FMNIST},
\item CIFAR-10,
32x32 RGB images; the three channels were averaged down to greyscale
\cite{CIFAR}.
\end{itemize}
All data sets are natively divided into training and test subsets.
In all cases, unless specified otherwise,
we constructed our training sets by using the first $P=8192$ elements of the 
training subset.
For computing the test errors, Eq.~(\ref{eq:errors_def}) below, we used the full test subsets.
In both training and test sets,
we binarised the classification task by using label $y=-1$ for odd classes and $y=1$ for even classes,
except when stated otherwise.
All inputs in our training sets (respectively, test sets)
were standardised by removing the mean
and dividing by the standard deviation, separately for each pixel $i$: 
$x_i \leadsto (x_i-\left<x_i\right>)/(\left<x_i^2\right>-\left<x_i\right>^2)^{1/2}$,
where the means $\left<x_i\right>$ and $\left<x_i^2\right>$ are computed on the 
training set (respectively, test set).

\paragraph{Projection onto the unit sphere}
The problem of linear separation of a set of points is projective, in the following sense.
Let us consider a linear separator identified by the vector
$w\in\mathbb R^H$. A dichotomy $f$ of the points $z^\mu\in\mathbb R^H$, $\mu=1,\ldots,P$,
can be defined by setting $f(z^\mu)=\mathrm{sign}(w\cdot z^\mu)$ for all $\mu$.
There is a large class of transformations of the points $z^\mu$ under which
the dichotomy is invariant.
In particular, rescaling each point by a (possibly different) positive factor,
$\tilde z^\mu = \lambda^\mu z^\mu$,
$\lambda^\mu > 0$, gives $f(\tilde z^\mu)=f(z^\mu)$.

This fact shows that projection onto the unit sphere of the internal representations
$h_t(x^\mu)$, as defined above Eq.~(\ref{eq:manifolds}),
does not affect the final linear readout.
We perform this normalisation step during evaluation of the metric quantities.
However, we do it for both shallow
and deep networks, even if the invariance does not hold for the latter.
Without this transformation, the non-monotonicity in the learning dynamics would be less evident
(see e.g.~\cite{SchillingMaier:2021}).

\paragraph{Definitions of the quantities measured}

The train and test errors were computed as
\begin{equation}{\label{eq:errors_def}}
\epsilon_{\mathrm{tr},\mathrm{test}} = 1-\frac{1}{\left|\mathcal T_{\mathrm{tr},\mathrm{test}}\right|}\sum_{x\in \mathcal T_{\mathrm{tr},\mathrm{test}}} \delta_{\hat y(x), y(x)},
\end{equation}
where $\mathcal T_{\mathrm{tr},\mathrm{test}}$ is the training or test set respectively,
and $y(x)=\pm 1$ is the label of $x$.

The squared gyration radii of the class manifolds, Eq.~(\ref{eq:manifolds}),
are defined as follows:
\begin{equation}
R^2_\pm(t) =
\frac{1}{2 n_\pm^2}
\sum_{x,y  \in \mathcal M_\pm(t)} \lVert x-y \rVert^2,
\end{equation}
where $n_+ = \left| \mathcal M_+(t) \right|$
is the number of elements with label $+1$
(and similarly for $n_-$).
The distance $D(t)$ between the centres of mass of
$\mathcal M_+(t)$ and $\mathcal M_-(t)$ is
\begin{equation}
D(t) = \left\lVert \frac{1}{n_+}\sum_{x \in \mathcal M_+(t)} x
- \frac{1}{n_-}\sum_{x \in \mathcal M_-(t)}  x\right\rVert.
\end{equation}

The inversion epoch $t_*$ is the epoch corresponding to the stationary value of
each metric quantity:
\begin{equation}
\begin{split}
t_*^{R_\pm} &= \underset{t}{\operatorname{argmin}} \; R_\pm(t),\\
t_*^{D} &= \underset{t}{\operatorname{argmax}} \; D(t)
\end{split}
\end{equation}
Operatively, we computed $t_*$ separately for $R_+$, $R_-$, and $D$;
then $\phi=\epsilon_\mathrm{tr}(t_*)$ was computed by averaging
the training errors corresponding to these values of $t_*$.
The values of $\phi$ reported are averages over 100 training runs.

\paragraph{Identity of stragglers}

To check that the identity of stragglers is conserved across
initialisations, we performed the following experiment.
We trained a two-layer neural network with 20 hidden units
(on MNIST with $P=8192$)
starting from two random initialisations.
For each of the two runs, $\alpha=0,1$, we identified the set $\mathcal S_\alpha(t_*)$ 
containing the misclassified elements of $\mathcal T$
at the inversion
epoch; in addition, we picked two random subsets $\widehat{\mathcal S}_\alpha\subset\mathcal T$,
such that $|\widehat{\mathcal S}_\alpha|=|\mathcal S_\alpha(t_*)|$.
We computed the numbers of common points
$M=| \mathcal S_0(t_*) \cap \mathcal S_1(t_*)|$ and
$\hat M=| \hat{\mathcal S}_0 \cap \hat{\mathcal S}_1|$.
The distributions of $M$ and $\hat M$, over 10k repetitions,
were peaked around $M\approx680$
and $\hat M\approx 70$, with standard deviations
$\sigma_M\approx 10$ and $\sigma_{\hat M}\approx 8$.
Comparing these numbers to the number of stragglers 
for this setting (around 800) shows that around 85\% of them are conserved,
as opposed to 9\% in the null model.

To quantify the stability of the stragglers,
or more in general of the sets $\mathcal S\left(t\left(\epsilon_\mathrm{tr}\right)\right)$,
we used the z-score
\begin{equation}
z = \frac{\left<M\right>-\left<\right.\!\hat M\!\left.\right>}{\sigma_M},
\end{equation}
where $\left<M\right>$ and $\left<\right.\!\hat M\!\left.\right>$
are the averages of $M$ and $\hat M$,
and $\sigma_M$ is the standard deviation of $M$,
obtained with a similar experiment as the one described above, but for different epochs $t(\epsilon_\mathrm{tr})$ instead of $t_*$.
The z-score for $\mathcal S\left(t\left(\epsilon_\mathrm{tr}\right)\right)$,
as a function of $\epsilon_\mathrm{tr}$, is plotted in Fig.~2d.

The z-score can be used to measure the conservation of the stragglers' identity
between different models.
To this aim, we used the same definition as above, and performed the two runs $\alpha=0,1$
on two possibly different architectures.
We obtained the following z-scores; in parentheses, the depths $L_\alpha$ and widths $H_\alpha$ of the two architectures,
$(L_0/H_0, L_1/H_1)$:
$z=50$ (2/20, 4/20),
$z=41$ (2/20, 8/20),
$z=40$ (4/20, 8/20).
By comparison, the same-architecture z-scores for the deeper models are
$z=51$ (4/20, 4/20) and
$z=45$ (8/20, 8/20).
The same method can be used to compare the identity of the stragglers
for two different shuffles of the training set, when training is performed
using stochastic gradient descent.
For a shallow network with 20 hidden units, and batch size 32,
we obtained $z\approx 11$.

\paragraph{Scaling with training set size}

A simple form, inspired by the theory of finite-size scaling in statistical physics \cite{Cardy:FSS},
is effective in capturing the dependence of the fraction of stragglers, $\phi$, on the size of
the training set $P$:
\begin{equation}\label{eq:fss}
\phi(P) = \phi_\infty \left[ 1- \left(\frac{P}{P_0}\right)^{-\gamma} + o\!\left(P^{-\gamma}\right) \right].
\end{equation}
Fits are performed by varying $\phi_\infty$, $P_0$, and $\gamma$.
The asymptotic values $\phi_\infty$ in Fig.~3(e)
were obtained by fitting
Eq.~(\ref{eq:fss}) to data with $P=4096, 8192, 16384, 32768$.


\section*{Data availability}
The datasets analysed during the current study are available in
public repositories;
links are in the corresponding publications \cite{MNIST,KMNIST,FMNIST,CIFAR}.

\section*{Code availability}
The code produced and used in the current study \cite{the_code}
is available on GitHub,
under the GNU General Public License, version 3 (GPL-3.0),
at
\url{https://github.com/marco-gherardi/stragglers}

\section*{Acknowledgements}
P.R~.acknowledges funding from the Fellini program under the H2020-MSCA-COFUND action, Grant Agreement No.~754496, INFN (IT).

\section*{Author contributions statement}
S.C.~and M.G.~discovered the stragglers.
M.G., M.O., and P.R.~conceived and designed the experiments.
L.C., S.C., M.G., and F.V.~performed the experiments.
All authors analysed the results and wrote the paper.
M.G.~and M.O.~supervised the analysis.
M.G.~coordinated the project.

\bibliographystyle{unsrt}
\bibliography{biblio}

\vspace{0.5cm}
\begin{center}
\includegraphics[width=8cm]{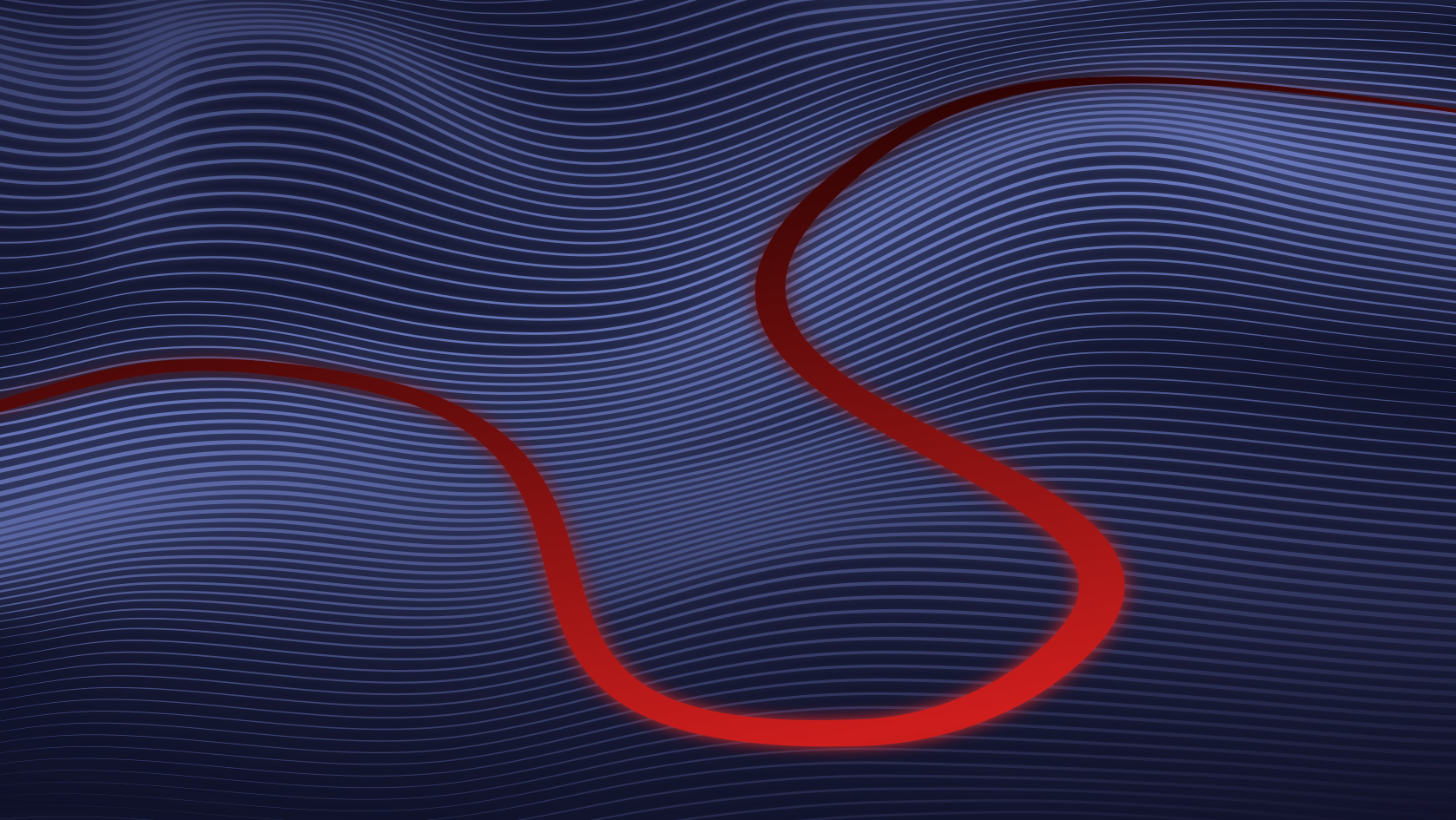}
\end{center}

\end{document}